%% file: Main_paper.tex
\documentclass[10pt,compsocconf,conference]{IEEEtran}
\IEEEoverridecommandlockouts
\newcommand{\ApproxSign}{\raise.17ex\hbox{$\scriptstyle\sim$}}

\usepackage[bookmarks=false]{hyperref}
\usepackage{cite}
\usepackage{lipsum}
\usepackage{multirow}
\usepackage{amsmath}
\usepackage{amssymb}
\usepackage{amsfonts}
\usepackage{algorithmic}
\usepackage{graphicx}
\usepackage{textcomp}
\usepackage{float}
\usepackage{blindtext}
\usepackage{xcolor}
\usepackage{caption}
\usepackage{array}
\usepackage{booktabs}
\newcolumntype{C}[1]{>{\centering\let\newline\\\arraybackslash\hspace{0pt}}m{#1}}

\def\BibTeX{{\rm B\kern-.05em{\sc i\kern-.025em b}\kern-.08em
    T\kern-.1667em\lower.7ex\hbox{E}\kern-.125emX}}

\newcommand{\Q}[1]{{}}

\newcommand\blfootnote[1]{%
  \begingroup
  \renewcommand\thefootnote{}\footnote{#1}%
  \addtocounter{footnote}{-1}%
  \endgroup
}
\usepackage{enumitem}
\setlist[enumerate]{leftmargin=*}
\usepackage{fancyhdr}

\hypersetup{
    colorlinks=true,
    linkcolor=red,
    filecolor=magenta,      
    urlcolor=blue,
}

\begin{document}

\title{E2GC: Energy-efficient  Group Convolution in Deep Neural Networks\\
{}
}
\author{\IEEEauthorblockN{Nandan Kumar Jha\textsuperscript{\dag}\thanks{$\dag$ Both authors contributed equally}, Rajat Saini\textsuperscript{\dag}, Subhrajit Nag, Sparsh Mittal}
\IEEEauthorblockA{
Department of Computer Science and Engineering, IIT Hyderabad, India\\ 
Email: \{cs17mtech11010, cs17mtech11002, cs17resch11006, sparsh\}@iith.ac.in
}}

\maketitle

\input{00Abstract}


\input{05Introduction}

\input{11Background}

\input{22RelatedWorks}

\input{33ProposedModel}
\input{44ExperimentalResults}

\input{55Discussion}

\input{66Conclusion}

{
\scriptsize
 \linespread{0.97}
\bibliographystyle{IEEEtran1}
\bibliography{References}
}

\end{document}

%% file: 00Abstract.tex
\begin{abstract}

The number of groups ($g$) in group convolution (GConv) is selected to boost the predictive performance of deep neural networks (DNNs) in a compute and parameter efficient manner. However, we show that naive selection of $g$ in GConv creates an imbalance between the computational complexity and degree of data reuse, which leads to suboptimal energy efficiency in DNNs. We devise an optimum group size model, which enables a balance between computational cost and data movement cost, thus, optimize the energy-efficiency of DNNs. Based on the insights from this model, we propose an ``energy-efficient group convolution'' (E2GC) module where, unlike the previous implementations of GConv, the group size ($G$) remains constant. Further, to demonstrate the efficacy of the E2GC module, we incorporate this module in the design of MobileNet-V1 and ResNeXt-50 and perform experiments on two GPUs, P100 and P4000. We show that,  at comparable computational complexity, DNNs with constant group size (E2GC) are more energy-efficient than DNNs with a fixed number of groups (F$g$GC). For example, on P100 GPU, the energy-efficiency of MobileNet-V1 and ResNeXt-50 is increased by 10.8\% and 4.73\% (respectively) when E2GC modules substitute the F$g$GC modules in both the DNNs. Furthermore, through our extensive experimentation with  ImageNet-1K and Food-101 image classification datasets, we show that the E2GC module enables a trade-off between generalization ability and representational power of DNN. Thus,  the predictive performance of DNNs can be optimized by selecting an appropriate $G$.  
The code and trained models are available at \url{https://github.com/iithcandle/E2GC-release.}

\end{abstract}

%% file: 05Introduction.tex
\section{Introduction}

The remarkable predictive performance of deep neural networks (DNNs) in various cognitive tasks such as image classification and object recognition   \cite{arXiv2017_Geirhos} led to its deployment on battery-driven systems running in self-driving cars, wearable devices, and robotics. Since these applications have stringent power constraints \cite{mittal2019SurveyJetson}, improving the energy efficiency of DNNs is of paramount importance to enable wide deployment in these domains. However, the high accuracy of DNNs comes at the cost of high computational complexity and large model size, which makes their energy consumption high \cite{2017_CVPR_Xie,He2016IdentityMI,ref102}. \blfootnote{This work was supported in part by Semiconductor Research Corporation (SRC) contract number 2018-IR-2852 and Science and Engineering Research Board (SERB), India, award number ECR/2017/000622.}

Recently, researchers have proposed several design techniques for reducing the number of computations and parameters by virtue of making DNNs compact \cite{Howard2017MobileNetsEC,2018_CVPR_Sandler,Zhang_2018_CVPR,Ma_2018_ECCV,2018_NIPS_Gao}. These design techniques include (1) low rank filters \cite{Ioannou_2016_ICLR}, (2) depthwise convolution (DWConv) \cite{Howard2017MobileNetsEC,2018_CVPR_Sandler}, (3) group convolution (GConv) \cite{2017_CVPR_Xie,2018_CVPR_Huang}, and (4) variants of DWConv and GConv \cite{2018_NIPS_Gao}. However, the effect of these design techniques on energy efficiency has not been systematically studied. The aforementioned design methods lead to {\em  different data layout and alter the memory access patterns}, which, in turn, affect the degree of data reuse and hence, the energy efficiency of DNNs. For example,  DWConv causes fragmented memory access and low data reuse \cite{Ma_2018_ECCV}, which leads to low resource utilization and low energy-efficiency on GPUs \cite{2019_Jha_VLSID,mittal2019SurveyDLGPU}. 

{\bf Motivation:} GConv enables compute and parameter efficient predictive performance \cite{2017_CVPR_Xie,Ioannou_2017_CVPR,2017_ICCV_Zhang,2018_CVPR_Xie,2018_BMVC_Sun,2018_CVPR_Huang,Zhang_2018_CVPR,2018_NIPS_Gao}. Increasing number of groups ($g$) in GConv reduces the number of computation, however, also reduces the data reuse. Lower data reuse results in high bandwidth pressure, and hence, it can outweigh the benefits of reduced computations in GConv. In Fig. \ref{fig:GConvEnergy}, energy consumed per frame ($EPF$), with various batch size ($B$), in MobileNet-V1 \cite{Howard2017MobileNetsEC} and ResNeXt-50 \cite{2017_CVPR_Xie} is shown (experimental setup is detailed in Section \ref{sec:ExperimentalResults}).

\input{01Fig_GConvEnergy}

Evidently, in both MobileNet-V1 and ResNeXt-50, $EPF$ first decreases with increasing $g$, reaches a minimum at $g=8$ or $g=16$, however, it starts increasing at higher $g$. This implies that at the lower $g$ decrease in computational cost is consequential. However, at the higher $g$, the effect of reduced computation is outweighed by lower data reuse, and hence, energy consumption starts increasing. {\em Unfortunately, previous works on GConv overlooked the interplay of $g$ and energy efficiency, which led to the energy-inefficient design of DNN}.

In this paper, we explore the architecture space of GConv with fixed and variable group sizes. We study the implications of different choices of group size on energy efficiency, and further, analyze the regularization effect {\em implicit} in GConv. To the best of our knowledge, this is the first work which explores the architecture space of GConv and shows the implicit trade-offs in GConv with constant group size (i.e., E2GC) that can optimize energy-efficiency as well as the predictive performance of DNNs. Our main {\bf contributions} are:

\noindent\textbf{1.} We show that naive selection of $g$ in GConv creates an imbalance between the number of computations and data reuse, which leads to suboptimal energy efficiency in DNNs.

\noindent\textbf{2.}  We devise an optimum group size model which enables a balance between computational cost and data movement cost and maximizes the energy efficiency of DNNs.

\noindent\textbf{3.}   Based on the insights from our optimum group size model, we propose an E2GC module and incorporated in the design of widely used compact DNN, MobileNet-V1, and a high-accuracy DNN, ResNeXt-50. 

\noindent\textbf{4.}  Through comprehensive experimentations on two GPUs, we show the efficacy of the E2GC modules. Further, we demonstrate the implicit trade-off in DNNs with the E2GC module, between generalization ability and representational power.

%% file: 01Fig_GConvEnergy.tex
\begin{figure}[htbp]\centering
\includegraphics[scale=0.16]{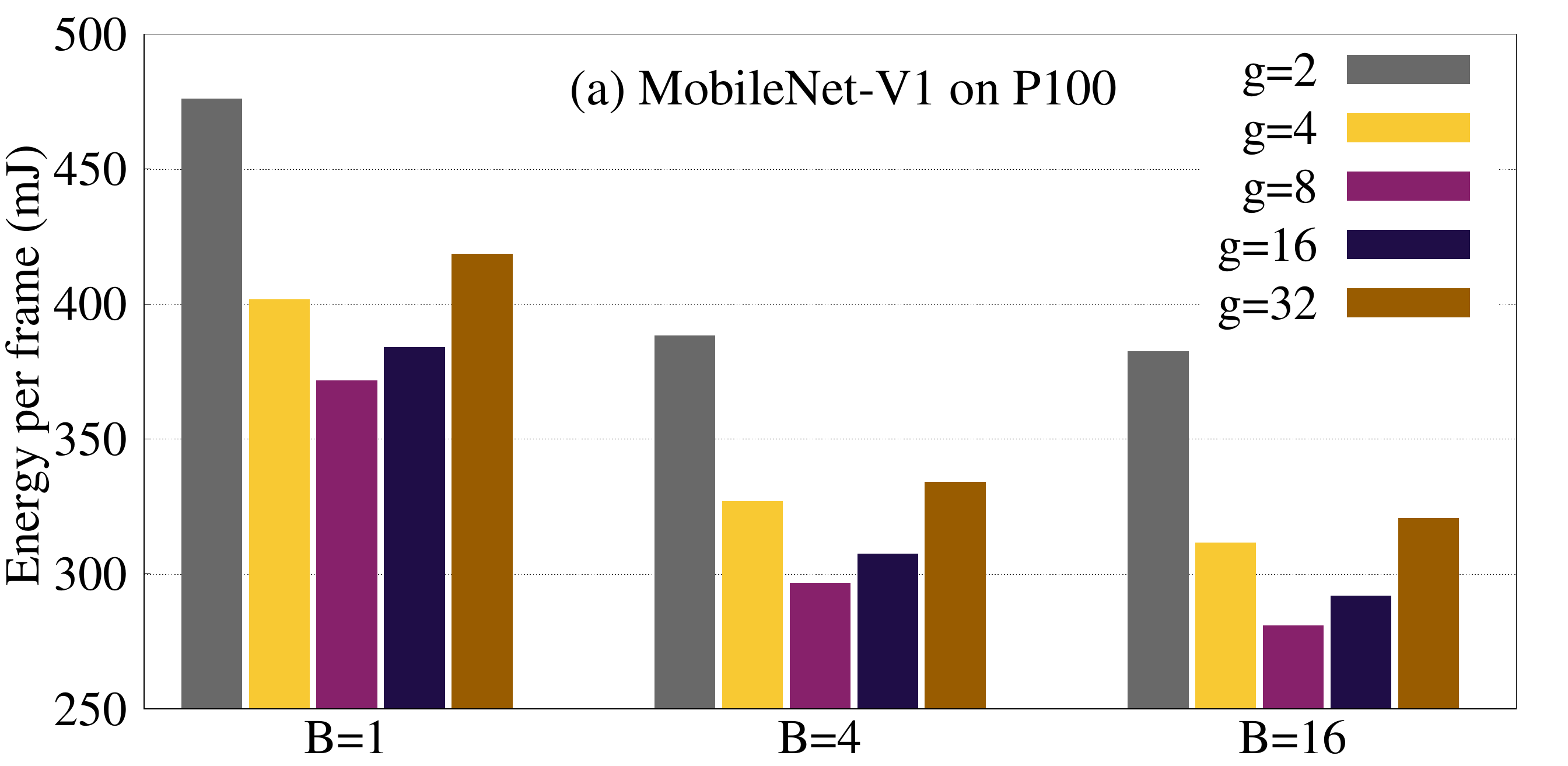}
\includegraphics[scale=0.16]{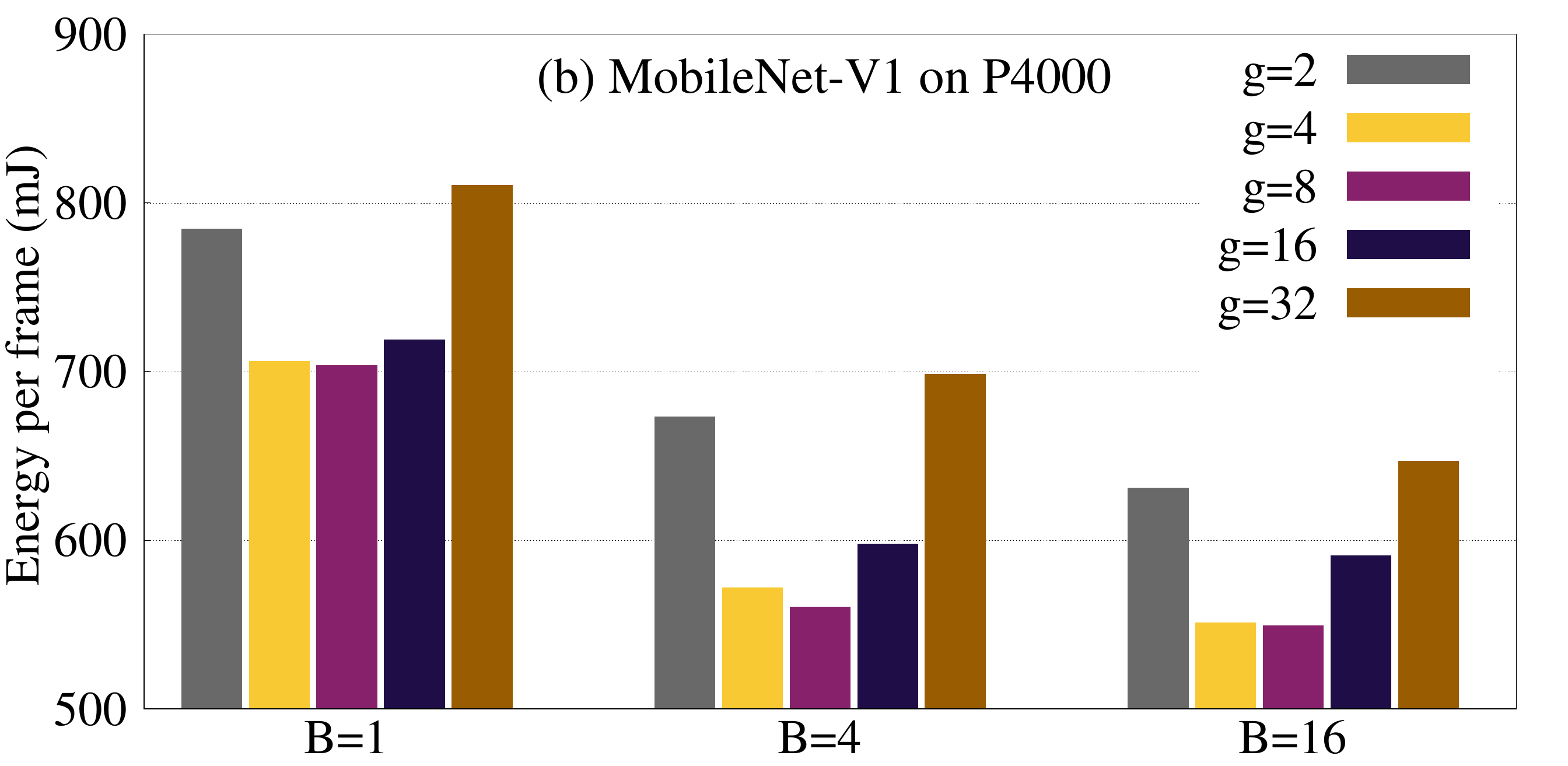} \\
\includegraphics[scale=0.16]{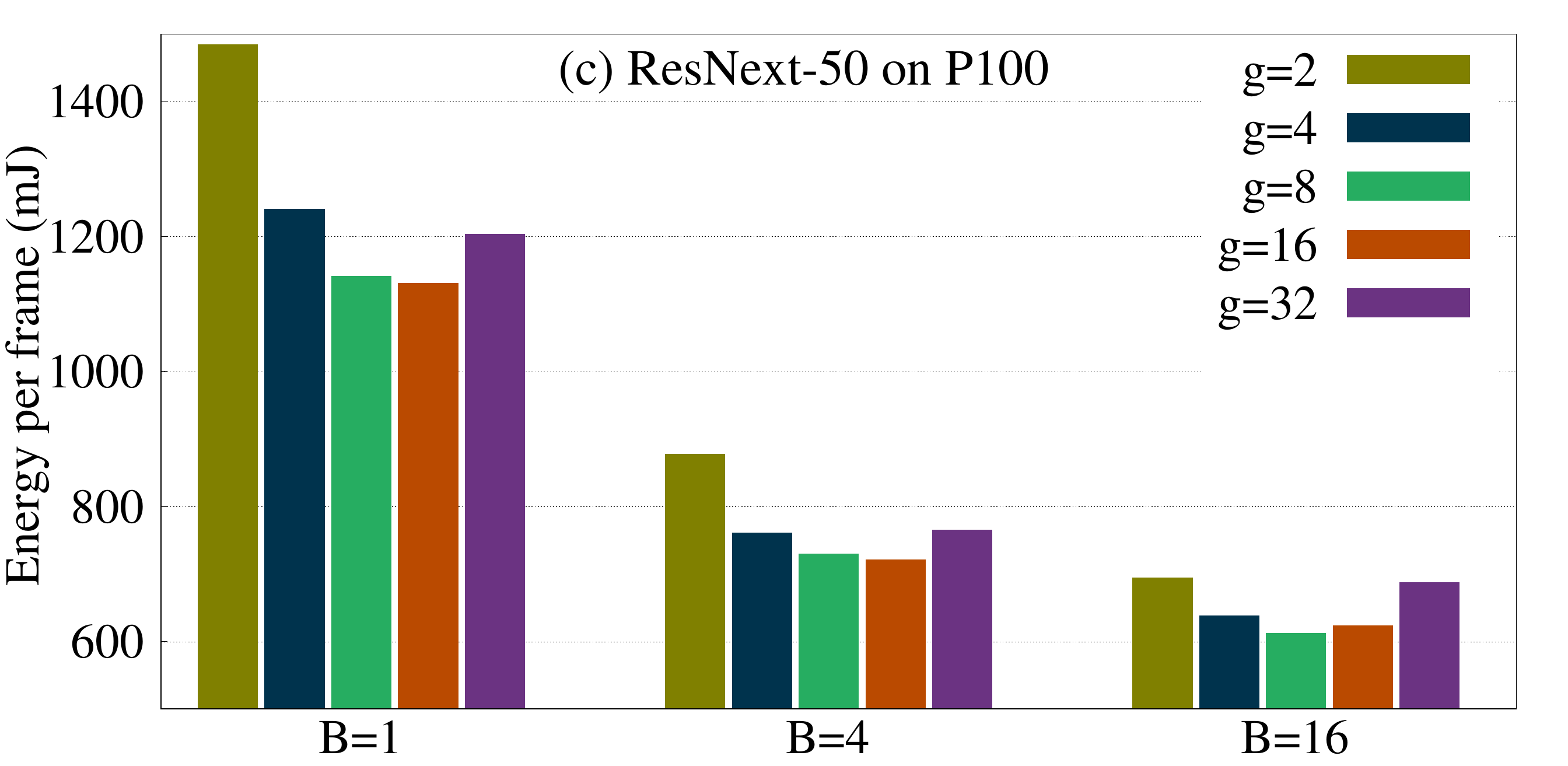}
\includegraphics[scale=0.16]{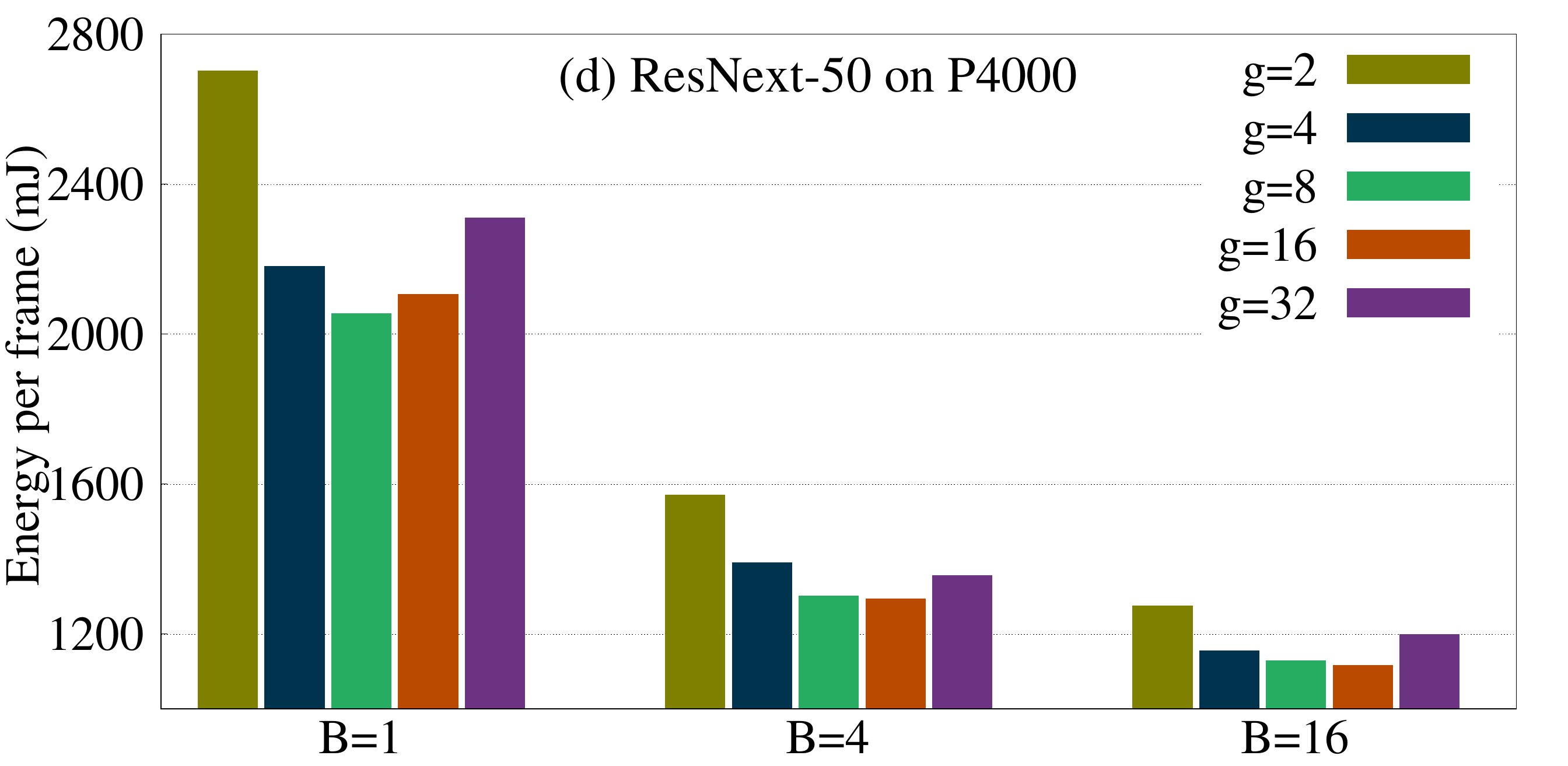}
\caption{$EPF$ for MobileNet-V1 (a, b) and ResNeXt-50 (c, d). }
\label{fig:GConvEnergy}
\end{figure}

%% file: 11Background.tex
\section{Background and related work}   \label{sec:Background}
In Table \ref{tab:Symbols}, we list the dimensions of filters, input feature maps (ifmaps), and output feature maps (ofmaps). The number of multiply-accumulation operations (MACs) calculated in this table correspond  to standard convolution operation. For simplicity, we assume that the spatial dimensions of ifmaps and ofmaps are equal.

\input{21Table_Symbols}

\textbf{Types of convolution and their compute efficiency:} In Fig. \ref{fig:convtypes}, we illustrate  different types of convolution used in DNN design. In standard convolution (SConv), filters are sparsely connected to the ifmaps and enable weight sharing. However,   each ofmap in SConv  is connected to every ifmap in a fully connected pattern (Fig. \ref{fig:convtypes}(a)). This connection pattern causes the term $n\times m$ in MAC calculation (Table \ref{tab:Symbols}) and makes SConv a compute-heavy operation.

To make convolution compute-efficient, GConv has been devised where only a group of ifmaps are connected to every ofmap (Fig. \ref{fig:convtypes}(b)). The number of MAC operations in each group is reduced by a factor of $g^2$ and overall computational complexity of GConv with $g$ groups reduced by a factor of $g$. The extreme case of GConv i.e., when $g$ = $m$ is known as DWConv \cite{Howard2017MobileNetsEC,2017_CVPR_Chollet,2018_CVPR_Sandler} and it reduces the computational complexity by a factor of $n$. In effect, GConv and DWConv achieve higher compute efficiency than other types of convolution. However, unlike the SConv, the ofmaps in GConv and DWConv receive the input information only from a fraction of ifmaps, and this reduces the predictive performance of networks \cite{Zhang_2018_CVPR,Ma_2018_ECCV}. To mitigate this issue, GConv and DWConv are often followed by pointwise ($1\times 1$) convolution (Fig. \ref{fig:convtypes}(d)) which blends the information across the ofmaps  produced by GConv(e.g., ResNeXt \cite{2017_CVPR_Xie}, MobileNet-V1 \cite{Howard2017MobileNetsEC}, MobileNet-V2 \cite{2018_CVPR_Sandler}).

\input{20Fig_ConvTypes}

\textbf{Sparsity and redundancy:} DNNs inherently possess a large amount of redundancy in terms of filter-weights, which stems from both the spatial and channel extent in filters \cite{NIPS_2013_Denil}. The redundancy in spatial extent is eliminated by using low-dimensional embeddings, such as $1\times 1$ filters \cite{2013_Lin_NiN},  and low-rank filters, such as $1\times 3$ and $3\times 1$ filters \cite{Ioannou_2016_ICLR}. These techniques reduce the DNN's computational complexity by lowering the value of $d_k\times d_k$ in MAC calculation (Table \ref{tab:Symbols}). However, the expensive $n\times m$ term remains unaffected, and hence, it is still compute-heavy. Some recent works such as MobileNet-V1 \cite{Howard2017MobileNetsEC} and EfficientNet \cite{2019_ICML_Tan} incorporated depth and width scaling to reduce the expensive $n\times m$ term. However, the fully connected pattern between ofmaps and ifmaps still exists.

In GConv (Fig. \ref{fig:convtypes}(b)) and DWConv  (Fig. \ref{fig:convtypes}(c)) only a group of ifmaps of  size $\frac{m}{g}$ interacts with each ofmap and hence,  break the fully connected pattern between ifmaps and ofmaps. In effect, GConv acts as coarse-grained and structured channel pruning where the redundant connections are selected randomly, i.e., irrespective of the importance of each ifmap. The channel sparsity introduced by the group of filters in GConv is conducive for better generalization and helps in achieving higher predictive performance \cite{2017_arXiv_Changpinyo}.   
{\em  Thus, GConv achieves higher compute efficiency along with structured sparsity, which enables better predictive performance when the information across the ofmaps are fused.}

%% file: 21Table_Symbols.tex
 \begin{table} [htbp] \small
\caption{Size of feature map is $h\times w$ and spatial dimension of filter is $d_k\times d_k$. Arith. = Arithmetic } \vspace{-0.1cm}
\label{tab:Symbols}\centering 
\resizebox{0.48\textwidth}{!}{
\begin{tabular}{ |c|c||c| } \hline
\textbf{Quantity (symbol)} & \textbf{Expression} & \textbf{Quantity (symbol)}  \\ \hline 
 \# MACs ($M_c$) &  $n\times m\times d_k^2\times h\times w $ & \# input fmaps  ($m$)  \\
 \# parameters ($P$) & $n\times m\times d_k^2$ & \# output fmaps  ($n$) \\
 \# activations ($A$) & $(n + m)\times h\times w$ & \# groups in GConv ($g$)  \\
 Arith. intensity ($AI$) & $M_c/(P+A)$ & Energy per frame ($EPF$)  \\
 \# channels in a group ($G$) & $m/g$ &  Group size in GConv ($G$)   \\
 \hline
\end{tabular}  }
\end{table}

%% file: 20Fig_ConvTypes.tex
\begin{figure}[htbp]\centering
\includegraphics[scale=0.35]{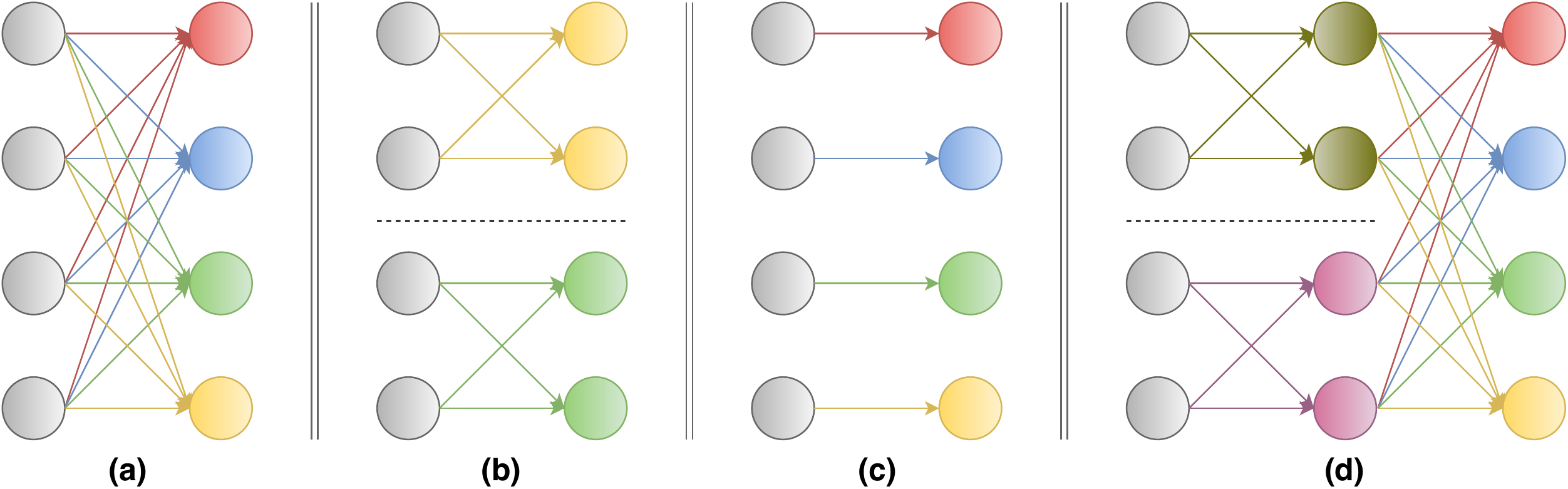}
\caption{Different types of convolution: (a) SConv. (b) GConv. (c) DWConv. (d) GConv followed by pointwise  convolution.}
\label{fig:convtypes}
\end{figure}

%% file: 22RelatedWorks.tex
{\bf DNNs with GConv/DWConv: } In AlexNet \cite{2012_NIPS_Krizhevsky}, GConv ($g$ = 2) was used to circumvent the limited memory issue in GPU and facilitate model parallelism. In ResNeXt \cite{2017_CVPR_Xie}, GConv ($g$ = 32) has been employed to boost the representational power of the network with reduced computations. CondenseNet \cite{2018_CVPR_Huang} employed learned GConv to remove the redundant connections in layer-wise dense connection patterns. IGCNet \cite{2017_ICCV_Zhang}, IGC-V2 \cite{2018_CVPR_Xie}, and IGC-V3 \cite{2018_BMVC_Sun} have deployed interleaved GConv comprised of primary and secondary GConv. Further, IGC-V2 and IGC-V3  use sparse and low-rank filters (respectively) to reduce the spatial redundancy in filters, which improved compute efficiency. ShuffleNetV1 \cite{Zhang_2018_CVPR} has GConv ($g$ = 3) to reduce the computational cost of pointwise convolution.  Deep Roots \cite{Ioannou_2017_CVPR} first performed a comparative analysis of GConv with increasing, decreasing, and constant $g$, then, incorporated a hierarchy of filters with improved compute efficiency and representational power.  ChannelNets \cite{2018_NIPS_Gao} deployed channel-wise GConv and DWConv and achieved high accuracy with reasonable computational complexity.  Changpinyo et al. \cite{2017_arXiv_Changpinyo} employed overlapping groups of filters and showed the effect of sparsity on the representational power of DNNs.

DWConv is a special case of GConv where $g$ = $m$ and enables high compute efficiency. DWConv followed by pointwise convolution employed in  XceptionNet \cite{2017_CVPR_Chollet}, MobileNet-V1  \cite{Howard2017MobileNetsEC}, MobileNet-V2 \cite{2018_CVPR_Sandler}, ShuffleNet-V1 \cite{Zhang_2018_CVPR}, and ShuffleNet-V2 \cite{Ma_2018_ECCV}. The  aforementioned works on GConv and DWConv chose  group size to boost the representational power in a compute-efficient manner. However, {\em they overlooked the interplay of $g$ and $G$ with energy efficiency of DNNs}. We propose E2GC module where $G$ is selected to maximize energy-efficiency.

%% file: 33ProposedModel.tex
\section{Proposed Method}  \label{sec:ProposedModel}

In this section, we first discuss the compute efficiency and reuse properties of GConv. We then devise an optimum group size model which strikes a balance between computational complexity and data reuse and optimized energy-efficiency. 

\subsection{GConv and  data reuse } \label{subsec:GConvDataReuse}

Let  $X$ $\in$ $\mathbb{R}^{m\times h\times w}$ be ifmaps transformed into ofmaps $Y$ $\in$ $\mathbb{R}^{n\times h\times w}$  by using SConv with filters $F$ $\in$ $\mathbb{R}^{n\times m\times d_k\times d_k}$ with appropriate padding and stride. In GConv, ifmaps and filters are divided into  $g$ disjoint groups, i.e. $X$ = \{$X_1$, $X_2$, ..., $X_g$\} and $F$ = \{$F_1$, $F_2$, ..., $F_g$\}, where each group in ifmaps has $G = \frac{m}{g}$ ifmaps and each group of filters has $\frac{n}{g}$ filters with $G$ channels i.e. $X_i$ $\in$ $\mathbb{R}^{\frac{m}{g} \times h\times w}$ and $F_i$ $\in$ $\mathbb{R}^{\frac{n}{g}\times \frac{m}{g}\times d_k\times d_k}$. This group division of ifmaps and filters affects the computational complexity and data (weights and activations) reuse behavior in DNN (Table \ref{tab:GConvMACsAI}). With increasing $g$, i.e. going from  SConv to DWConv (Fig. \ref{fig:ConvLine}), the value of $M_c$, $P$, and $AI$ decreasing while $A$ remains constant (Table \ref{tab:GConvMACsAI}). Decreasing $AI$ implies low data reuse i.e., increases memory access and bandwidth pressure. Note that in Fig. \ref{fig:ConvLine}, the number of memory access is  estimated ({\em theoretically}) as $\frac{1}{AI}$.

\input{30Table_MACandAI}

{\bf Intuition: } As shown in Fig. \ref{fig:ConvLine}, DWConv is most compute and parameter efficient but, due to the lower number of ifmaps per group ($G$ = 1), DWConv has lowest $AI$ and results into fragmented memory access. Hence, our intuition is that an increase in memory access would dwarf the benefit of reduced $M_c$ on the overall energy efficiency of DNNs. Also, the increase in $AI$ is slower compared to the increase in $M_c$ at higher $G$ (Fig. \ref{fig:ConvLine}). These observations motivate us to find the optimal value of  $G$ (or $g$) where the increase in $M_c$ would be balanced by the increase in data reuse ($AI$); thus, energy-efficiency would be maximized. Further, since, changing $G$ alters the degree of sparsity and regularization strength in DNNs,  we also need to investigate the interplay of $G$ and predictive performance of DNNs. We aim to achieve a balance between energy-efficiency, storage overhead ($P$), and predictive performance of DNNs. For this purpose, we now devise an optimum group size model which would help DNN designers in finding the optimal value of $g$.

\input{28Fig_ConvMACsLine}

\subsection{Optimum group size modeling} \label{subsec:EnergyModel}

As discussed in the previous section, a decrease in $M_c$ is associated with an increase in memory access, i.e., both are {\em tightly coupled}. The decrease in $M_c$ reduces energy consumption while increasing memory access leads to higher energy consumption. Moreover, the energy consumed in off-chip memory access is orders of magnitude higher than that in an arithmetic operation \cite{horowitz2014computing}. To account for this disparity in energy consumption, we take a variable $\beta \in (0,1)$ in Eq. \ref{eqn:MACandAI}. Further, the number of memory access to different levels in the memory hierarchy depends on (1) the implicit data reuse available in DNN (e.g., DWConv has lower $AI$ compared to SConv),  (2) architecture of memory hierarchy which includes number of levels and storage capacity of memory levels in hierarchy,  and (3) memory management policies such as implicit vs. explicit, coupled vs. decoupled, and placement and update policy \cite{2019_ASPLOS_Pellauer}. To approximate these factors in energy efficiency, we take a variable $\alpha \in (0,1]$ in Eq.  \ref{eqn:MACandAI}. 

{\bf Physical significance of $\alpha$ and $\beta$}:
The value of $\alpha$ closer to zero implies better on-chip data reuse, whereas a value of  $\beta$ closer to zero indicates a larger disparity between the energy consumption of arithmetic operation and memory access. Note that both $\alpha$ and $\beta$ are platform-dependent, which can vary between 0 and 1.

We aim to find the optimal value of $g$, where the increase in memory access does not outweigh the decrease in $M_c$. To balance these factors, we have the following equations.

\begin{gather} 
M_c \times \Big(\frac{\alpha}{AI}\Big)^{\beta} = const ; \; \; \text{where} \; \; AI = \frac{M_c}{A + P} \nonumber \\ \implies \label{eqn:MACandAI}
(M_c)^{1-\beta} \times (A + P)^{\beta} = const \times \alpha^{-\beta}
\end{gather}

In Eq. \ref{eqn:MACandAI}, $const$ is a constant to strike a balance between memory access cost and computational cost. Now, in Eq. \ref{eqn:MACandAI}, we take a variable $\gamma$ = $const \times \alpha^{-\beta}$. Since we intend to find the optimal value of $g$, and $A$ does not depend on $g$ (Table \ref{tab:GConvMACsAI}), the Eq. \ref{eqn:MACandAI} is transformed into following equations.

\begin{gather}
(M_c)^{1-\beta} \times P^{\beta} = \gamma \nonumber \\ 
\text{After substituting $M_c$ and $P$ from Table II, we have} \nonumber \\ 
\Bigg(\frac{m\times n\times h\times w\times d_k^2}{g}\Bigg)^{1-\beta}\times \Bigg(\frac{m\times n\times d_k^2}{g}\Bigg)^{\beta} = \gamma \nonumber \\  \implies \label{eqn:GConvFn}
g = \frac{m\times n\times d_k^2\times (h\times w)^{1-\beta}}{\gamma}  
\end{gather}

The optimal value of $g$, which enables a balance between computational cost and memory access cost  depends on the filter dimensions; $m$, $n$, and $d_k$ and ofmap size; $h$ and $w$ (Eq. \ref{eqn:GConvFn}). Therefore, {\em keeping $g$ constant in all the layers of a DNN, as done in previous implementations of GConv ($g$ = 32 in \cite{2017_CVPR_Xie}), is not conducive from energy-efficiency standpoint}.

The dimensions  $m$, $n$, $h$, and $w$ vary across the layers in DNNs. However, in most state-of-the-art DNNs, except the first convolution layer,  the spatial size of the filter ($d_k\times d_k$) does not vary across the layers.  For example, in MobileNet-V1, MobileNet-V2, and ResNeXt, almost all layers have only $3\times 3$  and $1\times 1$  filters. For such DNNs, on a fixed hardware platform ($\gamma$ will be constant) 
\begin{gather} 
g  \propto m\times n\times (h\times w)^{(1-\beta)} \nonumber \\
\text{Let $h$ = $w$  = $d_f$ and $\beta$ = 0.5 \text{( Since, $\beta \in (0,1]$ )}} \nonumber \\ \implies \label{eqn:GConvProp}
g  \propto m\times n\times d_f
\end{gather}

To avoid the representational bottleneck in DNNs \cite{Szegedy_2016_CVPR}, the number of filters ($n$) increased by $2\times$ when each spatial dimension of ofmap decreased by $0.5\times$. For example, in ResNeXt and MobileNet-V1, $d_f\times d_f$ becomes $\frac{d_f}{2}\times \frac{d_f}{2}$ in $3\times 3$ GConv/DWConv layer and $n$ becomes 2$n$ in the successive $1\times 1$ layer. For such DNNs, Eq. \ref{eqn:GConvProp} transformed into following.  
\begin{gather}
\label{eqn:Gproportional}
g \propto m \implies G = const_1 \; \; (\text{since, $G$ = $\frac{m}{g}$ })
\end{gather}

In above equation, $const_1$ is a constant which is fixed for a particular DNN on a particular hardware platform. Based on the findings and insights from Eq. \ref{eqn:MACandAI}, Eq. \ref{eqn:GConvFn}, Eq. \ref{eqn:GConvProp}, and Eq. \ref{eqn:Gproportional}, we propose energy-efficient GConv (E2GC) module where GConv is followed by a pointwise convolution layer. In the GConv of E2GC modules  value of $g$ is selected in proportion to the value of $m$, i.e., number of channels ($G$) remains constant in all the layers of DNN (Fig. \ref{fig:E2GC}(a)).

\input{32Fig_GConv}

Since $G$ is constant in Gconv of E2GC modules, $g$ increases in deeper layers (Fig. \ref{fig:E2GC}(a)). However, in conventional GConv, $g$ remains constant and $G$  increases in deeper layers (Fig. \ref{fig:E2GC}(b)), e.g., in ResNeXt ($g$ = 32) and ShuffleNet-V1 ($g$ = 3). {\bf We termed this conventional GConv ($g$ is constant) followed by a pointwise convolution layer as F$g$GC module} (Fig. \ref{fig:E2GC}(b)). Note that in both  E2GC and F$g$GC modules,  $1\times 1$ layer  blends the information from all the groups in ofmaps and avoids the drop in predictive performance of DNN. 

%% file: 30Table_MACandAI.tex
\begin{table} [htbp] 
\caption{$M_c$, $P$, $A$ (ifmaps + ofmaps) and $AI$ in GConv. } \vspace{-0.15cm}
\label{tab:GConvMACsAI}\centering 
\resizebox{0.48\textwidth}{!}{
\begin{tabular}{ |c|c|c|c|} \hline
 $M_c$ & $P$ & $A$ (ifmaps + ofmaps) & $AI$ \\ \hline 
  $\frac{n\times m\times  h\times w\times d_k^2}{g}$ & $\frac{n\times m\times d_k^2}{g}$ & $(n + m)\times h\times w $ & $\frac{n\times m\times  h\times w\times d_k^2 }{n\times m\times d_k^2 + g\times (n + m)\times h\times w }$ \\ \hline
\end{tabular} } 
\end{table}

%% file: 28Fig_ConvMACsLine.tex
\begin{figure}[htbp]\centering
\includegraphics[scale=0.335]{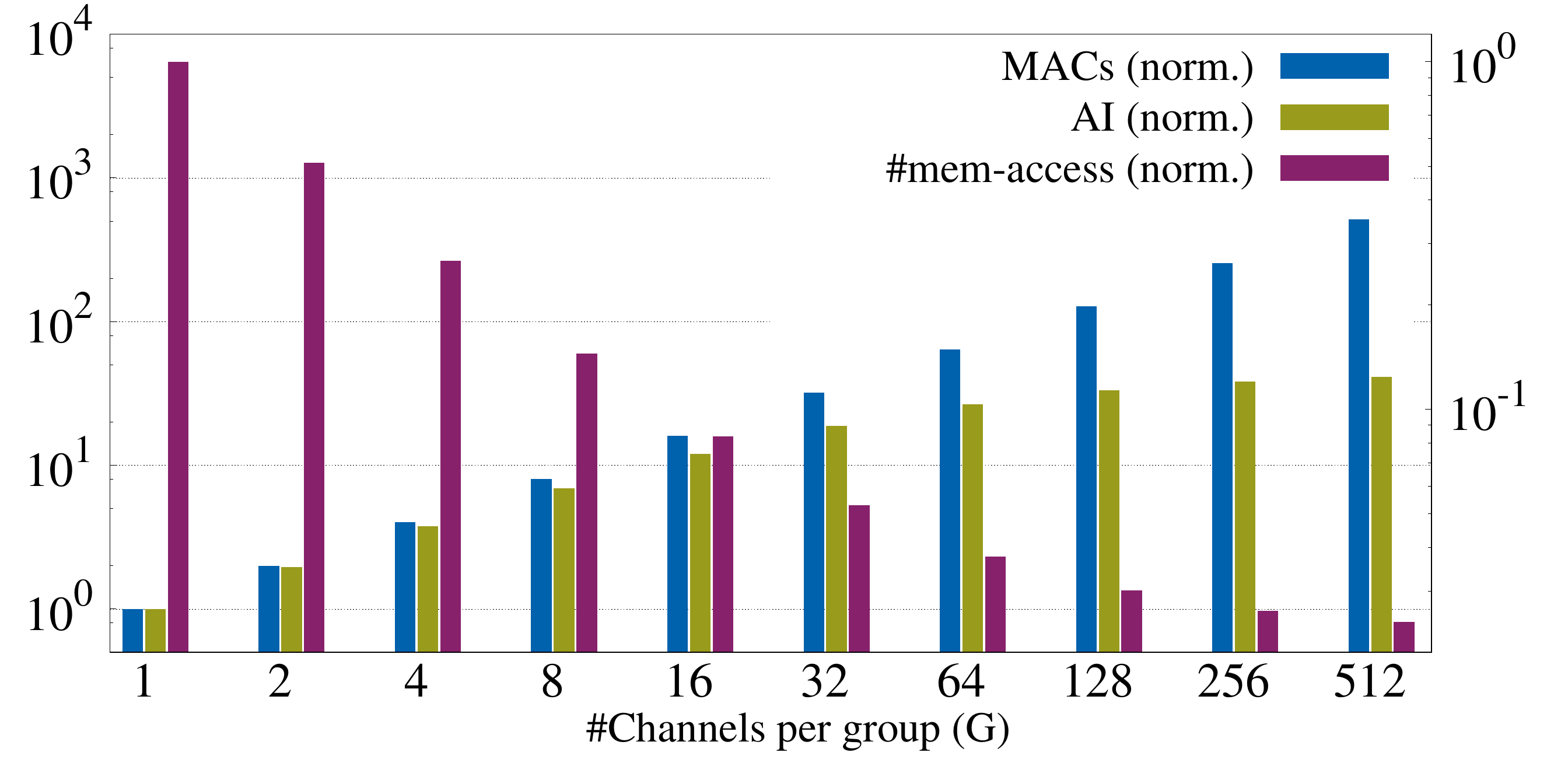}
\includegraphics[scale=0.65]{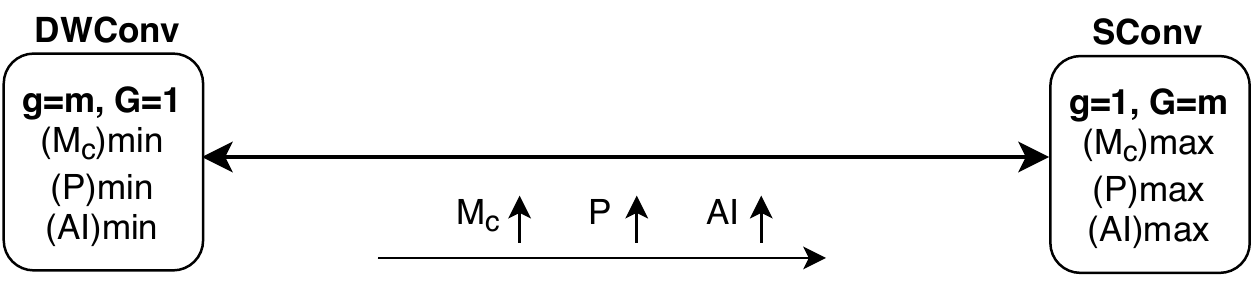}
\caption{$M_c$ and  $AI$ (left vertical axis), and number of memory access (right vertical axis) are normalized to that in DWConv with $n$ = $m$ = 512, $d_k\times d_k$ = $3\times 3$, and $h\times w$ = $14\times 14$. }
\label{fig:ConvLine}
\end{figure}

%% file: 32Fig_GConv.tex
\begin{figure}[htbp]\centering
\includegraphics[scale=0.45]{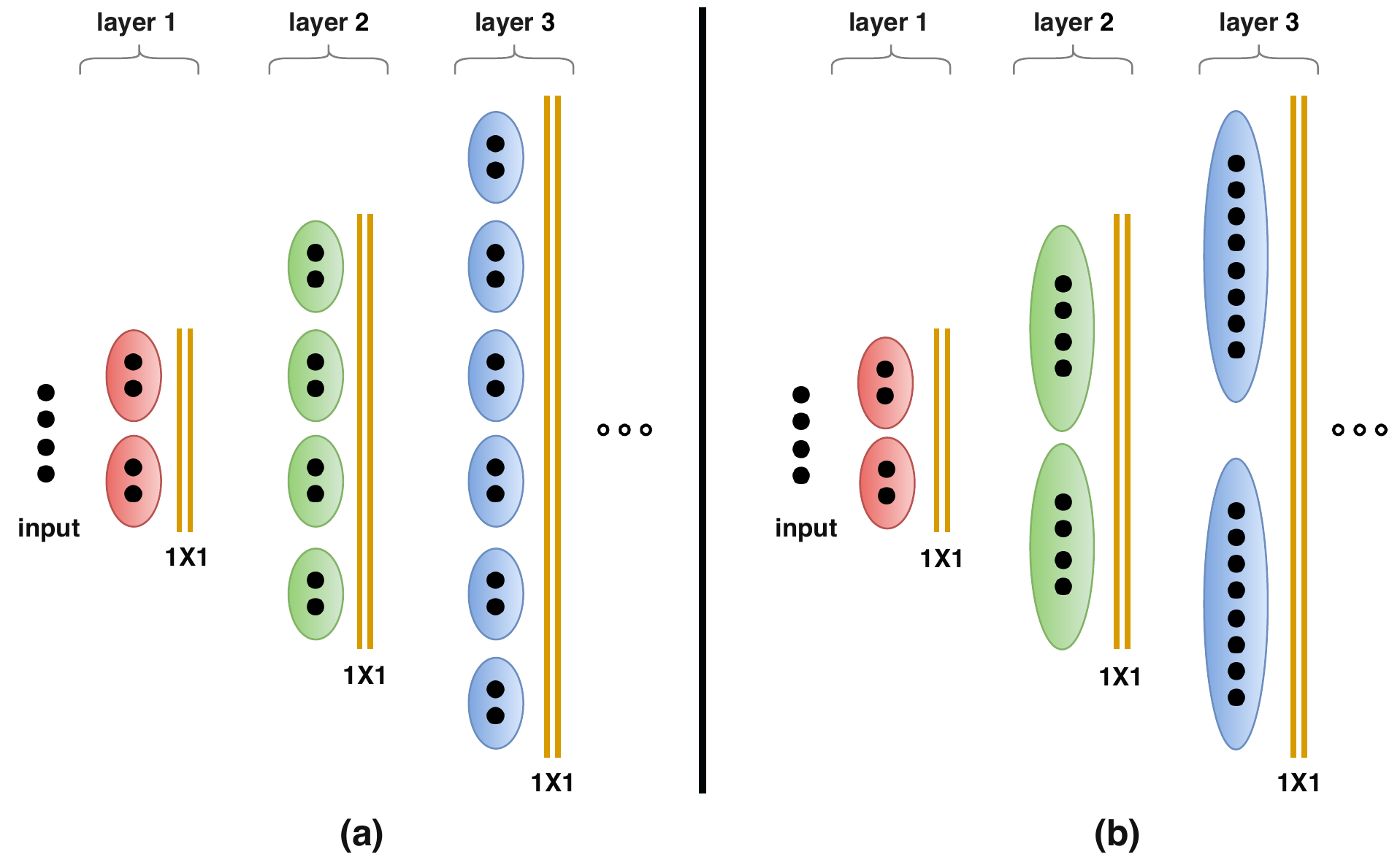}
\caption{(a) Proposed E2GC module where $G$ remains constant,  (b) conventional F$g$GC module where $g$ remains constant in all the convolution layers of DNNs with GConv.}
\label{fig:E2GC}
\end{figure}

%% file: 44ExperimentalResults.tex
\section{Experimental Results }  \label{sec:ExperimentalResults}

\textbf{Dataset.} 
We perform our experiments on two image classification datasets, ImageNet-1K and Food-101 \cite{2014_ECCV_Bossard}. The former consists of  1.2M training and 50K validation samples categorized among 1000 classes, whereas the latter consists of 75,750 training and 25,250 validation samples categorized among 101 classes. Food-101 is a fine-grained and challenging image classification dataset which has high variations in local information \cite{2014_ECCV_Bossard} and also, the training images are noisy. We use a single center crop of size 224$\times$224 for both datasets. 

\textbf{Experimental setup:} 
We evaluate the efficacy of  E2GC module on ResNeXt-50 (32 x 4d), which employed GConv ($g$=32) followed by $1\times 1$ layer, and MobileNet-V1 which employed DWConv ($G$=1) followed by $1\times 1$ layer.  We replace ``GConv/DWConv followed by $1\times 1$ layer" with E2GC/F$g$GC modules in both the DNNs. For training, we used SGD optimizer with 0.9 Nesterov momentum and 4e-5 weight decay rate with 0.1 initial learning rate, which is reduced to $\frac{1}{10}$th after every 30 epochs. All the Top-1 accuracies reported in this paper are with a single crop on a single model. We used {\tt nvidia-smi} tool for measuring the power consumption on GPUs.  For training, we use PyTorch deep learning framework. To validate the consistency of our findings, we measure $EPF$  with multiple $B$ as a larger $B$ leads to better utilization of the compute resources in the GPUs. We measure $EPF$ for one forward and one backward pass successively.

\input{41Table_MobileNetV1}

\input{42Table_ResNext}

{\bf Salient features of E2GC modules:} 
In Section \ref{subsec:EnergyModel}, we {\em theoretically} showed that, under certain conditions, keeping $G$ constant in all convolution layers is amenable for better energy efficiency. Whereas, fixing  $g$ in GConv is not conducive for higher energy efficiency. We validate our optimum group size model on  P100 and P4000 GPU and compare the performance of E2GC and F$g$GC modules. The results are shown in Table \ref{tab:MV1Results} and Table \ref{tab:ResNxtResults}. We make the following key observations.

{\em Observation 1:}
For both ResNeXt-50 and MobileNet-V1, at a comparable value of $M_c$ or $P$, E2GC provides higher energy efficiency than F$g$GC. For example, MobileNet-V1 with E2GC modules and $G$ = 32, is more energy-efficient than both MobileNet-V1 with F$g$GC modules and  $g$ = 8, and $g$ = 16, which have comparable $M_c$ and $P$ respectively (Table \ref{tab:MV1Results}). Similarly, in Table \ref{tab:ResNxtResults}, ResNeXt-50 with F$g$GC modules and $g$ = 32, has lower energy-efficiency than both ResNeXt-50  with E2GC modules and $G$ = 6,  and $G$ = 16, which have comparable $M_c$ and $P$ respectively. \\
{\em Observation 2:} 
At a comparable $M_c$, E2GC module yields lower  $P$ than the F$g$GC modules.  For example, MobileNet-V1 with F$g$GC modules and $g$ = 32, has comparable $M_c$ but higher $P$ compared to MobileNet-V1 with E2GC modules and $G$ = 8 (Table \ref{tab:MV1Results}). Similarly, ResNeXt-50 with F$g$GC modules and $g$ = 8,  has comparable $M_c$ but higher $P$ compared to ResNeXt-50 with E2GC modules and $G$ = 8.  Hence employing E2GC modules make DNN parameter-efficient.\\
{\em Observation 3: } 
Compared to ResNext-50 with F$g$GC modules and  $g$ = 16, ResNext-50 with E2GC modules and $G$ = 16  has lower energy-efficiency   at smaller $B$, however, at $B=16$ the latter has higher energy-efficiency than the former (Table \ref{tab:ResNxtResults}). This is because of the better resource utilization at higher $B$. \\ 
{\em Observation 4: } 
Because of better data reuse at  smaller values of $g$ in  DNNs with F$g$GC modules, it has higher energy efficiency  despite having very high $M_c$.  However, this results in very high $P$.
For example, MobileNet-V1 with $g$ = 2,  has $\approx 5\times$  number of MACs but lower $EPF$  than  MobileNet-V1 with $G$ = 1 (Table \ref{tab:MV1Results}). However, the former has $4\times$ number of parameters compared to the latter.

{\bf Optimal value of $g$ in DNNs with F$g$GC modules: } The $EPF$ initially decreases with increasing $g$, reaches a minimum at $g=8$ (or $g=16$), and then again starts increasing for both MobileNet-V1 and ResNeXt-50.  That is,  at higher $g$, the decrease in computational cost is inconsequential and dwarfed by the increase in memory access cost, whereas, at lower $g$, the increase in computational cost outweighed the benefit of higher data reuse.

{\bf Energy comparison with baseline models:} The baseline models for MobileNet-V1 and ResNeXt-50 families (i.e., ResNeXt-50 and MobileNet-V1 with E2GC/F$g$GC modules) are highlighted in Table \ref{tab:MV1Results} and Table \ref{tab:ResNxtResults}, respectively. In both the families, $EPF$ is {\em lowest} when  E2GC modules with $G=32$ are employed in the networks. In fact, MobileNet-V1 with E2GC modules and $G=32$ is 55.4\% and 65.4\%, whereas, ResNeXt-50 with E2GC modules and $G=32$ is 9.4\% and 13.1\% more  energy-efficient than their baseline model on P100 and P4000 GPUs (respectively).

These observations and results substantiate our claims that: (1) naive selection of $G$ or $g$ in GConv results in suboptimal energy efficiency, and (2) at comparable $M_c$, DNN with E2GC modules is more energy-efficient than a DNN with F$g$GC modules. In other words, {\em GConv with constant $G$ is more energy-efficient than GConv with fixed $g$}.

%% file: 41Table_MobileNetV1.tex
\begin{table*}[htbp]
\caption{ $EPF$ and classification accuracy of MobileNet-V1 employing group convolution with different group settings.  }
\label{tab:MV1Results}
\centering
\begin{tabular}{|l|c|c|c|c|c|c|c|c|c|c|} \hline
\multirow{2}{*}{MobileNet-V1 } & \multirow{2}{*}{ Params ($\times 10^6$) } & \multirow{2}{*}{ MACs ($\times 10^6$)}  & \multicolumn{3}{c|}{ EPF on P100 (millijoule)} & \multicolumn{3}{c|}{ EPF on P4000 (millijoule)} & \multicolumn{2}{c|}{ Accuracy (Top-1 $\%$) } \\ 
& & & B=1 & B=4 & B=16 & B=1 & B=4 & B=16 & ImageNet-1K & Food-101 \\ \hline
{\bf E2GC ($G$=1)}\cite{Howard2017MobileNetsEC} & 4.20 & 568.74 & 689 & 630 & 613 & 1607 & 1448 & 1373 & {\bf 70.65} & 79.96 \\
E2GC ($G$=2) & 4.25 & 586.13 & 538 & 482 & 450 & 1169 & 1014 & 983 & {\bf 72.76} & 80.48 \\
E2GC ($G$=4) & 4.34 & 620.90 & 421 & 357 & 330 & 1005 & 846 & 816 & 72.24 & 80.23 \\
E2GC ($G$=8) & 4.52 & 690.44 & 373 & 316 & 302 & 780 & 630 & 592 & 71.85 & 80.03 \\
E2GC ($G$=16) & 4.87 & 829.53 & 370 & 284 & 277 & 689 & 551 & 507 & 71.18 & 79.79 \\
E2GC ($G$=32) & 5.59 & 1107.71 & 363 & 281 & 265 & 648 & 501 & 480 & 72.76 & 79.69 \\ \hline
F$g$GC ($g$=2) & 16.72 & 2690.06 & 476 & 388 & 383 & 785 & 673 & 631 & 74.43 & 78.60 \\ 
F$g$GC ($g$=4) & 10.44 & 1620.71 & 402 & 327 & 312 & 706 & 572 & 551 & 73.59 & 79.06 \\
F$g$GC ($g$=8) & 7.30 & 1086.03 & 372 & 297 & 281 & 704 & 561 & 550 & 73.34 & 79.60 \\
F$g$GC ($g$=16) & 5.73 & 818.69 & 384 & 308 & 292 & 719 & 598 & 591 & 72.60 & 80.03 \\
F$g$GC ($g$=32) & 4.95 & 685.02 & 418 & 334 & 321 & 811 & 698 & 647 & 72.20 & 80.13 \\ \hline
\end{tabular}
\end{table*}

%% file: 42Table_ResNext.tex
\begin{table*}[htbp]
\caption{ $EPF$ and classification accuracy of ResNeXt-50 employing group convolution with different group settings. }
\label{tab:ResNxtResults}
\centering
\begin{tabular}{|l|c|c|c|c|c|c|c|c|c|c|} \hline
\multirow{2}{*}{ ResNeXt-50} & \multirow{2}{*}{ Params ($\times 10^6$) } & \multirow{2}{*}{ MACs ($\times 10^9$)} & \multicolumn{3}{c|}{ EPF on P100 (millijoule)} & \multicolumn{3}{c|}{ EPF on P4000 (millijoule) } & \multicolumn{2}{c|}{ Accuracy (Top-1 $\%$) } \\ 
& & & B=1 & B=4 & B=16 & B=1 & B=4 & B=16 & ImageNet-1K & Food-101 \\ \hline
E2GC ($G$=1) & 23.61 & 4.02 & 2185 & 1248 & 712 & 5780 & 2371 & 1434 & {\bf 74.43} & 82.62 \\
E2GC ($G$=2) & 23.68 & 4.05 & 1921 & 1000 & 678 & 4661 & 2109 & 1347 & {\bf 77.04} & 82.78 \\
E2GC ($G$=4) & 23.82 & 4.10 & 1476 & 804 & 667 & 3431 & 1674 & 1292 & 77.22 & 82.72 \\
E2GC ($G$=8) & 24.09 & 4.20 & 1162 & 742 & 631 & 2337 & 1318 & 1152 & 77.60 & 82.10 \\
E2GC ($G$=16) & 24.63 & 4.40 & 1132 & 722 & 619 & 2220 & 1251 & 1080 & 77.64 & 82.03 \\
E2GC ($G$=32) & 25.72 & 4.80 & 1088 & 694 & 597 & 2063 & 1189 & 1049 & 77.45 & 81.46 \\ \hline
F$g$GC ($g$=2) & 46.18 & 7.70 & 1485 & 878 & 695 & 2702 & 1572 & 1276 & 77.58 & 78.03 \\ 
F$g$GC ($g$=4) & 34.86 & 5.85 & 1241 & 761 & 638 & 2181 & 1391 & 1155 & 77.62 & 78.16 \\
F$g$GC ($g$=8) & 29.20 & 4.92 & 1142 & 730 & 612 & 2055 & 1302 & 1129 & 77.64 & 78.88 \\
F$g$GC ($g$=16) & 26.37 & 4.46 & 1131 & 722 & 624 & 2106 & 1295 & 1117 & 77.72 & 80.00 \\
{\bf F$g$GC ($g$=32)} \cite{2017_CVPR_Xie} & 24.96 & 4.23 & 1204 & 766 & 687 & 2310 & 1356 & 1200 & 77.80 & 80.05\\  \hline
\end{tabular}
\end{table*}

%% file: 55Discussion.tex
\section{Insights from classification accuracies} \label{sec:Discussion}

In the previous section, we demonstrate how the choices of $g$ and $G$ in E2GC and F$g$GC module (respectively) affect the energy-efficiency of DNNs. In this section, we explain how the choices of $g$ and $G$ in  E2GC and F$g$GC module affects the generalization ability and representational power of DNNs. 

{\bf Implicit regularization in GConv:} 
GConv acts as structured DropConnect \cite{PMLR_2013_Wan} where the drop units are a contiguous region in ifmaps, i.e., connections in channel extent (Fig.  \ref{fig:convtypes}(b)). However, DropConnect \cite{PMLR_2013_Wan} works well on fully connected layers but perform poorly on convolution layers where features are spatially correlated \cite{NIPS_2018_Ghiasi}. Since GConv drops a contiguous region in ifmaps, it can block the flow of a few semantic information from one layer to the next layer (similar to DropBlock \cite{NIPS_2018_Ghiasi}) and forces the network to learn simple co-adaptations. Hence, the discriminative regions in networks learn robust features which enable a better generalization.

{\bf Trade-off between generalization ability and representational power in GConv:} 
The limited number of channels in filter groups of GConv reduces the filter co-dependence and force the network to learn with limited filter-dependence \cite{Ioannou_2017_CVPR}. Limited co-dependence improves the generalization ability of network similar to \cite{PMLR_2013_Wan,NIPS_2018_Ghiasi}. However, filter groups with limited filter-dependence may not be able to capture enough variations of complex latent concepts, thus, reduces the level of abstraction and hence representational power of network \cite{2013_Lin_NiN}. Since  DNNs with the E2GC modules have a fixed number of connections in each group across all the layers, it introduces better-structured regularization compared to DNNs with F$g$GC modules where $G$ is increasing with $m$.

{\bf E2GC and ImageNet-1K:} 
When value of $G$ in MobileNet-V1 and ResNeXt-50 with E2GC modules increases from $G=1$ to $G=2$, {\em surprisingly, there is a substantial gain}, 3.5\% in ResNeXt-50 (Table \ref{tab:ResNxtResults}) and 3\% in MobileNet-V1 (Tables \ref{tab:MV1Results}), in Top-1 accuracy. Here the benefit of stronger regularization, introduced by only one channel-wise connection in GConv with $G=1$, is dwarfed by very limited filter-dependence in each group.  Further, increasing the value of $G$ beyond two results in a decreased Top-1 accuracy of MobileNet-V1 on ImageNet-1K. Since an increase in $G$ leads to a higher number of channel-wise connections in each group, GConv with a higher $G$, introduces weaker regularization. However, at $G=32$, filter groups capture enough variations of complex concepts and outweigh the effect of weaker regularization. Thus, at $G=32$, there is a significant boost in the Top-1 accuracy of MobileNet-V1 on ImageNet-1K. 

Compared to MobileNet-V1, ResNeXt-50 is a deeper and wider model (more layers with higher $m$ and $n$).  Hence, (1) it is more sensitive to representational power at a smaller $G$ in the E2GC module, and (2) compared to  MobileNet-V1,  ResNeXt-50 needs stronger regularization to improve the generalization ability. This leads to an increase in classification accuracy as $G$ is increasing beyond two; however, at $G$ = 32, the effect of weaker regularization outweighs the gain in representational power. Thus, there is a drop in Top-1 accuracy.

{\bf F$g$GC and ImageNet-1K:}
On increasing $g$ from 2 to 32 in GConv of F$g$GC modules used in MobileNet-V1, the  Top-1 accuracy decreases. The increase in $g$ implies a decrease in $G$ and stronger regularization, but variations of complex concepts that each group can capture also decrease. Altogether, the effect of lowering representational power outweighs the effect of stronger regularization in GConv of F$g$GC modules used in MobileNet-V1. In ResNeXt-50, the effect of regularization, introduced by GConv in F$g$GC modules, dominates and hence increase in  $g$, leads to an increase in Top-1 accuracy on ImageNet-1K (Tables \ref{tab:MV1Results} and  \ref{tab:ResNxtResults}).

{\bf Results on Food-101 dataset:}
Unlike ImageNet-1K, the Food-101 dataset has low inter-class variation but substantially high variation in local information. Hence, the effect of regularization dominates the effect of representational power in both,  DNNs with E2GC  and F$g$GC modules. Hence, for both DNNs,  Top-1 accuracy decreases with rising  $G$ and increases with rising $g$ (Tables \ref{tab:MV1Results} and Table \ref{tab:ResNxtResults}).

From the above analysis on Top-1 accuracy on two image classification datasets, it is evident that employing E2GC modules in DNNs enables a trade-off between generalization ability and representational power of DNNs. However, due to the varying $G$ in the F$g$GC module, DNNs with F$g$GC modules can tune either generalization ability or representational power through varying the value of $g$.

%% file: 66Conclusion.tex
\section{Conclusion}
We proposed energy-efficient group convolution, which balances the computational complexity and data movement cost in DNNs. We showed that GConv with constant group size (E2GC) is more energy-efficient than GConv with a fixed number of groups (F$g$GC). Also, by using an appropriate $G$, the predictive performance of the network can be optimized. 

